\documentclass[letterpaper, 10 pt, conference]{ieeeconf}  
\IEEEoverridecommandlockouts                              
\usepackage{graphics}    
\usepackage{times}       
\usepackage{amsmath}     
\usepackage{amssymb}     
\usepackage{graphicx}
\usepackage[font=small]{caption} 
\usepackage{tabularx}
\usepackage{makecell}
\usepackage{multirow}
\usepackage{hhline}
\usepackage[ruled,noline,noend]{algorithm2e}
\usepackage{tikz}
\usetikzlibrary{backgrounds, fit, shapes.geometric, positioning}
\usepackage[noend]{algpseudocode}

\def\figref#1{Fig.~\ref{#1}}
\def\tabref#1{Tab.~\ref{#1}}
\def\eqref#1{Eq.~(\ref{#1})}

\newcommand\etal{\emph{et al.}}

\newcolumntype{Y}{>{\centering\arraybackslash}X}
\title{\LARGE \bf The Pitfall of More Powerful Autoencoders in Lidar-Based Navigation 
}

\author{Christopher Gebauer \and Maren Bennewitz
  \thanks{All authors are with the Humanoid Robots Lab, University of Bonn, Germany. This work has partly been supported by the German Research Foundation under Germany’s Excellence Strategy, EXC-2070 - 390732324 (PhenoRob).
  }%
}

\begin{document}
\maketitle
\thispagestyle{empty} 
\pagestyle{empty}

\begin{abstract} 
    The benefit of pretrained autoencoders for reinforcement learning in comparison to training on raw observations is already known~\cite{yarats19}. In this paper, we address the generation of a compact and information-rich state representation. 
In particular, we train a variational autoencoder for 2D-lidar scans to use its latent state for reinforcement learning of navigation tasks. To achieve high reconstruction power of our autoencoding pipeline, we propose an - in the context of autoencoding 2D-lidar scans - novel preprocessing into a local binary occupancy image. This has no additional requirements, neither self-localization nor robust mapping, and therefore can be applied in any setting and easily transferred from simulation in real-world.
In a second stage, we show the usage of the compact state representation generated by our autoencoding pipeline in a simplistic navigation task and expose the pitfall that increased reconstruction power will always lead to an improved performance.
We implemented our approach in python using tensorflow. Our datasets are simulated with pybullet as well as recorded using a slamtec rplidar A3. 
The experiments show the significantly improved reconstruction capabilities of our approach for 2D-lidar scans w.r.t. the state of the art. However, as we demonstrate in the experiments the impact on reinforcement learning in lidar-based navigation tasks is non-predictable when improving the latent state representation generated by an autoencoding pipeline. This is surprising and needs to be taken into account during the process of optimizing a pretrained autoencoder for reinforcement learning tasks. \end{abstract} 

\section{Introduction}
\label{sec:intro}
In recent years, the amount of research addressing mobile robotics keep increasing and concurrently the number of thinkable applications for such platforms rises. However, one of the most fundamental tasks, to safely navigate within the environment, still leaves us with open difficulties. Modern systems consider the nearby surroundings based on sensor data, and even react reasonably to moving obstacles if their motion is mathematically describable~\cite{missura19}. Deep reinforcement learning on the other side is promising for transferring an existing approach to a new mobile platform or into an unknown environment. When the training procedure is setup correctly, the agent can be easily retrained for new tasks while still outperforming the classical approaches in terms of quality of the navigation policy~\cite{regier20},~\cite{sergeant15} 

\begin{figure}[t]
  \centering
  \begin{tikzpicture}[
    tri/.style={isosceles triangle,draw, inner sep=1mm},
    left/.style = {shape border rotate=180}
  ]
      \node[draw, rectangle, text centered, inner sep=0mm]                      (m0)    at (0, 4.23)       {\includegraphics[width=0.8\linewidth]{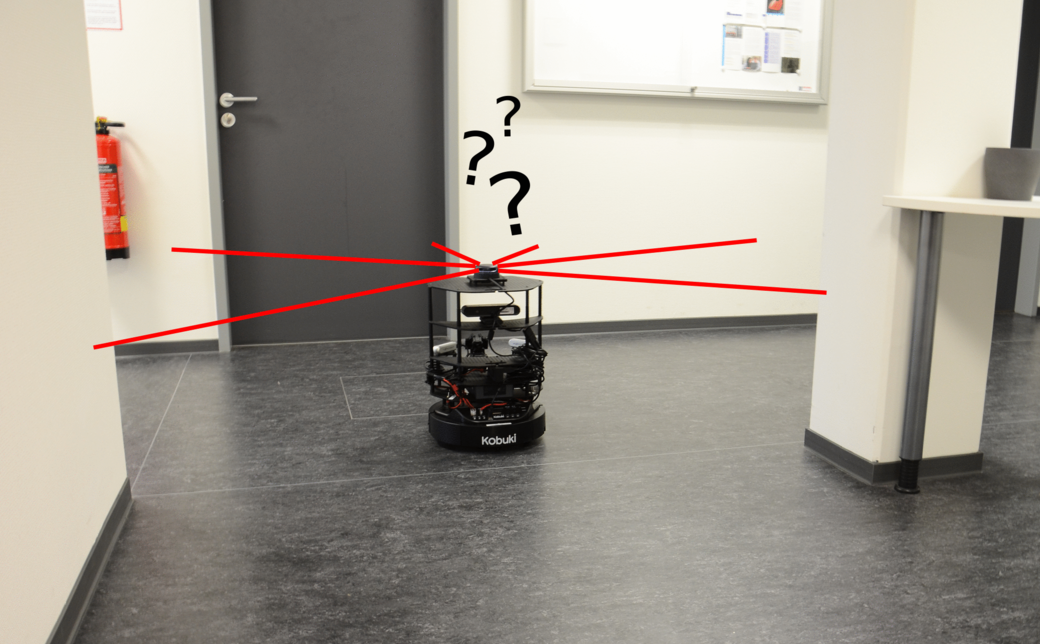}};
      \node[rectangle, text centered, inner sep=0mm]                      (m1)    at (0,-0.0)       {\includegraphics[width=0.8\linewidth]{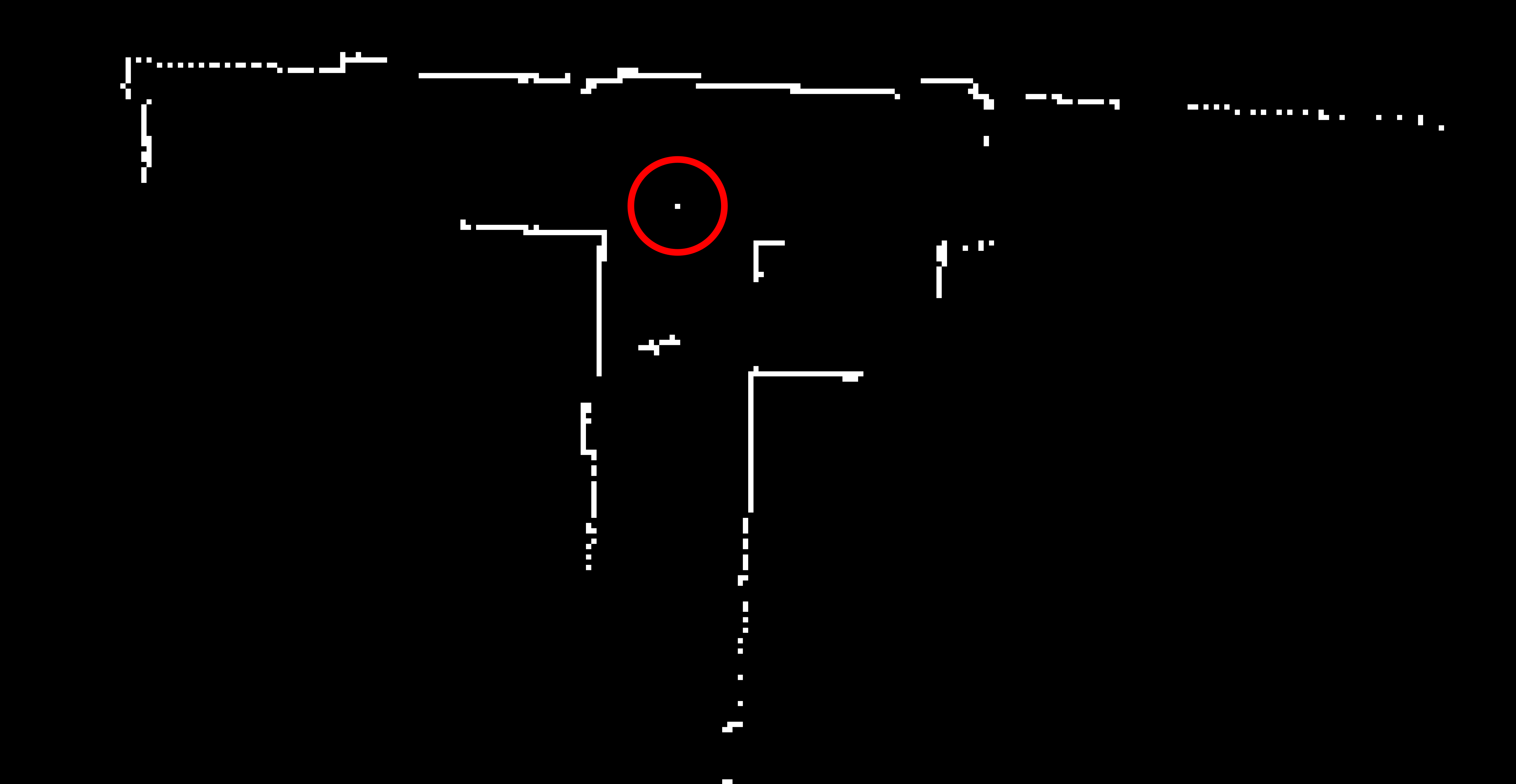}};
      \node[draw, white, rectangle, rounded corners, text centered, text=white, text width=20mm] (m1_) at (2.0, -0.8) {binary occupancy image};
      \node[draw, rectangle, rounded corners, text height=15mm, minimum height=20mm]    (m2)    at (0,-3.07)      {\includegraphics[width=0.77\linewidth]{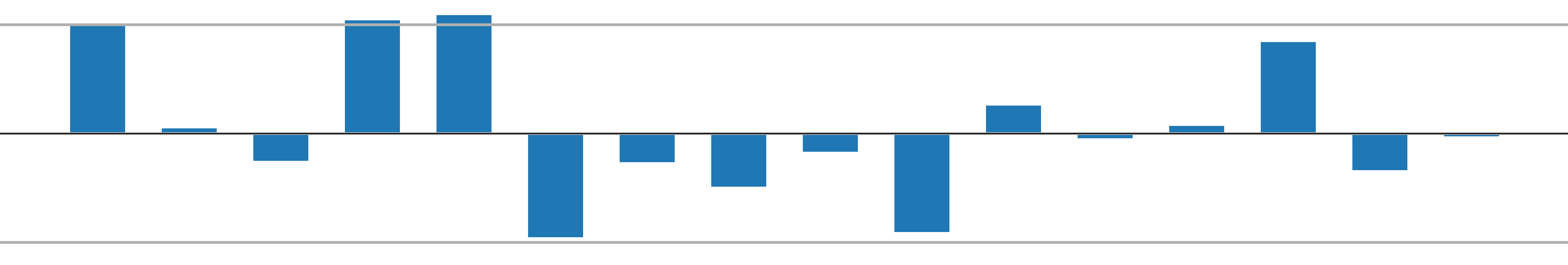}};
      \node[rectangle, text centered] (m2_) at (0.0, -2.45) {latent state};
      \node[draw, rectangle, text centered, rounded corners, minimum width=0.8\linewidth, minimum height=8mm]                    (m3)    at (0.0,-4.8)       {Policy};

      \draw[->, line width=0.4mm] (m0) -- (m1);
      \draw[->, line width=0.4mm] (m1) -- (m2);
      \draw[->, line width=0.4mm] (m2) -- (m3);
      \draw[->, line width=0.4mm] (m3.west)  -- ++ (-0.35, 0.0) |- (m0.west);
  \end{tikzpicture}

  \caption{An overview of the information flow in a reinforcement learning setup. The agent observes the current state of the robot~(top image), given by the 2D-lidar scan. Our proposed pipeline preprocesses the observation into a binary occupancy image, which is then encoded by a pretrained variational autoencoder into a latent state representation. Based on this state representation a policy is optimized and outputs an optimal action that is send to the robot. The position of the robot in the binary occupancy image is marked with a red circle.}
  \label{fig:motivation}
\end{figure}
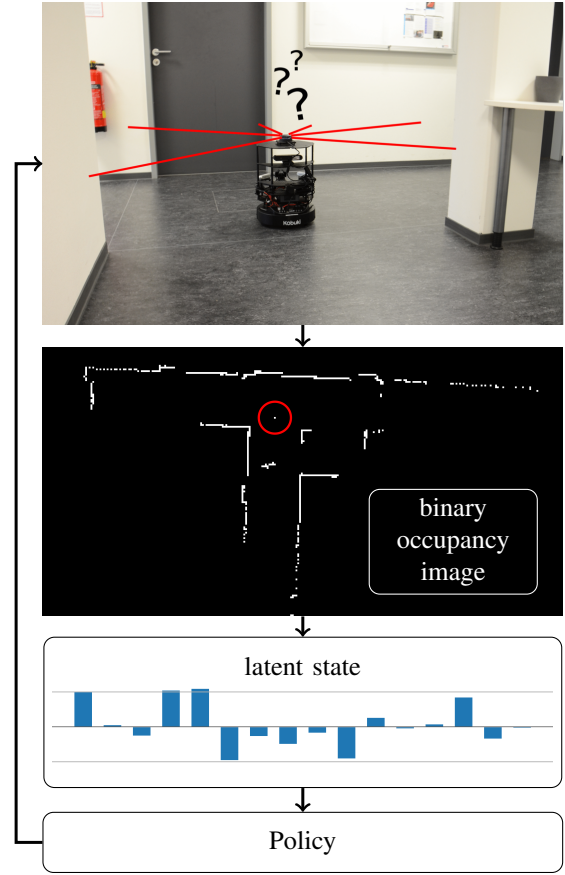 
The general idea of reinforcement learning~(RL) inherits the interaction of the agent with its environment by receiving the current observation and choosing an optimal action according to it, see also~\figref{fig:motivation}. However, Yarats~\etal~\cite{yarats19} show that RL approaches based on raw observations lead to low performance regarding both, the resulting quality of the agent as well as the training speed in clock time. To overcome this, usually powerful representations are trained in advance~\cite{finn16}, or concurrently with strong auxiliary losses~\cite{jaderberg16}. Both approaches are based on an autoencoder~(AE) architecture. The general idea of an AE is to compress the input into a latent space, where the target is to losslessly reconstruct the input based on this latent space. The increase in performance by applying a pretrained autoencoder to a RL task leads to the assumption, that improving the performance of the autoencoder will also improve the performance of the RL agent. However, we will show in our work that this assumption needs to be treated with care.

For indoor environments a common choice for mobile robots to observe their own state within the environment are 2D-lidar sensors, due to their robustness and low acquisition cost. For that reason we propose a, to our best knowledge, novel autoencoding pipeline for 2D-lidar scans. Our pipeline consists of a preprocessing into a local occupancy image and a variational autoencoder~(VAE) that compresses the preprocessed input into a probability distribution over the latent space. We, hereby focus on a compact latent space that can be further used for RL tasks in robot navigation, see also~\figref{fig:motivation}. As we show in our experiments, our pipeline has a significantly improved reconstruction power compared to an autoencoder based on the encoder architecture of Pfeiffer~\etal~\cite{pfeiffer16}. However, we also demonstrate the pitfall that in navigation tasks greater reconstruction power always leads to an improved performance of the resulting agent. To overcome this, the improvement of an input pipeline for RL should never be considered as a detached problem from the actual RL task. Rather it is recommended to alternate between a pipeline-specific metric and the performance of the downstream RL agent.

\section{Related Work}
\label{sec:related}
In recent years autoencoding lidar scans has been used for different purposes, but has drawn rather less attention for explicit representation learning. However, the performance of autoencoder in lidar-based navigation tasks has already been proven by Sergeant~\etal~\cite{sergeant15}, but not in a RL setup. The focus of their work was to reconstruct lidar data and previous actions into future actions alongside with the lidar scan itself. The autoencoder was trained using restricted Boltzmann machines and data recorded of a human expert, which results in an agent being able to follow a long corridor with the same behavior the human expert did.

Korthals~\etal~\cite{korthals19} apply a VAE for multi-modal sensor fusion, which means encoding a 2D-lidar scan and an RGB image of the identical scene into one latent space. The focus rather lays on the sensor fusion than on the further usage of the generated state representation. \mbox{Lundell~\etal~\cite{lundell18}} use an AE to reconstruct the full boundary shape of sparse obstacles such as tables, where only the legs are detected within the scan. This approach makes use of the effect that high discontinuities are averaged by a 1D convolutional layer, which we will discuss later on. The purpose is to support the navigation using the reconstructed data, while the latent space itself is not further considered.

Schlichting~\etal~\cite{schlichting18} train an AE for dimensionality reduction of a 2D-lidar scan and further consider the latent dimension for localization purpose. Matching scans are computed with a k-means clustering algorithm directly in the latent space. The context of RL is not addressed, however, the usability of an autoencoding pipeline for 2D-lidar scans is demonstrated. Wakita~\etal~\cite{wakita18} apply a VAE for map-construction and self-localization. A special focus lays on the averaging effect of the 1D convolution mentioned above and the resulting step edge in the reconstructed scan. The authors noted that this behavior has a tremendous effect on the reconstruction quality and needs to be avoided. The authors address this by training an additional network to distinguish between true edges of the environment and step edges generated by reconstructing the lidar data. As we are mainly interested in the latent space and not in the reconstructed data such post-processing is not reasonable in our application.

Gaccia~\etal~\cite{gaccia19} show that 2D representations of lidar data have a great advantage when processing in an autoencoding pipeline. However, they denoise 3D-lidar data and therefore the preprocessing is not directly applicable to a 2D-lidar scanner. We propose a local occupancy image, which can be directly computed based on the lidar data without additional need of any localization. Such computation has already been considered by Sax~\etal~\cite{sax19} but not in local fashion. The work rather focuses on supporting an exploratory policy by rewarding the agent for discovering unseen global grid cells, while the actual state received by the RL agent is based on RGB images. Narasimhan~\etal~\cite{narasimhan20} use a similar local occupancy map, but require a global pose and predict the map using seq2seq networks given the past and a RGB image instead of local sensor data.

We noticed that step edges introduced by Wakita~\etal~\cite{wakita18} only occur when using the 1D array of the raw range measurements as direct input. Therefore, we propose an - in the context of autoencoding 2D-lidar scans - novel preprocessing to overcome the problem of step edges. This results in a significantly improved reconstruction power of the autoencoding pipeline.  
\section{Background}
\label{sec:back}
In this section we briefly outline the required background of variational autoencoder~(VAE)~\cite{kingma13}. Autoencoders~(AE) in general consist of an encoder or recognition model~$q$, which compresses the data~$x$ into a latent space~$z \in \mathbb{R}^{k}$. The dimension or compactness of the latent space is given by the hyperparameter~$k$. The second part of an AE is the decoder or generative model~$p$, which returns the reconstructed input~$x'$ given the latent space. The target is to minimize the error between~$x$ and~$x'$. 

While the original concept of an AE is deterministic, Kingma~\etal~\cite{kingma13} applied amortized variational inference to propose a stochastic AE, known as variational autoencoder. The recognition model~$q(z|x)$ is changed to return a distribution over the latent space, which captures uncertainty. The generative model~$p(x'|z)$ returns a distribution over the reconstructed input given the latent space. Commonly, both models are parametrized by a neural network. The target is formulated by maximizing a variational lower bound on the marginalized log-likelihood of the generative model 
\begin{eqnarray}
\label{eq:elbo}
log \, p(x) \ge \mathbb{E}_{q} \left[ log \, p(x|z) \right] - \beta \cdot D_{\mathrm{KL}} \left( q(z|x)||p(z) \right).
\end{eqnarray}
The scaling factor~$\beta$ is introduced by Higgins~\etal~\cite{higgins17} to further support the simplicity of the latent space. For the original VAE it is assumed to be 1. Intuitively, the first term on the right in the above equation can be interpreted as a reconstruction loss. The second term is a penalty to prevent that the recognition model is too deterministic and concurrently maintain its simplicity. More precisely, the reconstruction term maximizes the log-likelihood of the generative model under the expectation of the recognition model. This can be directly computed, while the expectation can be estimated with one sample if the batch size is large enough~\cite{kingma13}. The penalty for simplicity is a distance measure, represented by the Kullback-Leibler~(KL) divergence, between the recognition model and a prior~$p(z)$. The prior is commonly defined by a diagonal standard Gaussian to emphasize the recognition model to be simple.  
\section{Proposed Autoencoding Pipeline}
\label{sec:main}

\begin{figure}[t]
  \centering
  \begin{tikzpicture}[
    tri/.style={isosceles triangle,draw, inner sep=1mm},
    left/.style = {shape border rotate=180}
  ]
      \node[draw, rectangle, text centered, minimum width=1, minimum height=0.5]                    (r)    at (0.0,0)       {\includegraphics[trim=3.0cm 3.0cm 3.0cm 3.0cm, width=0.1\linewidth]{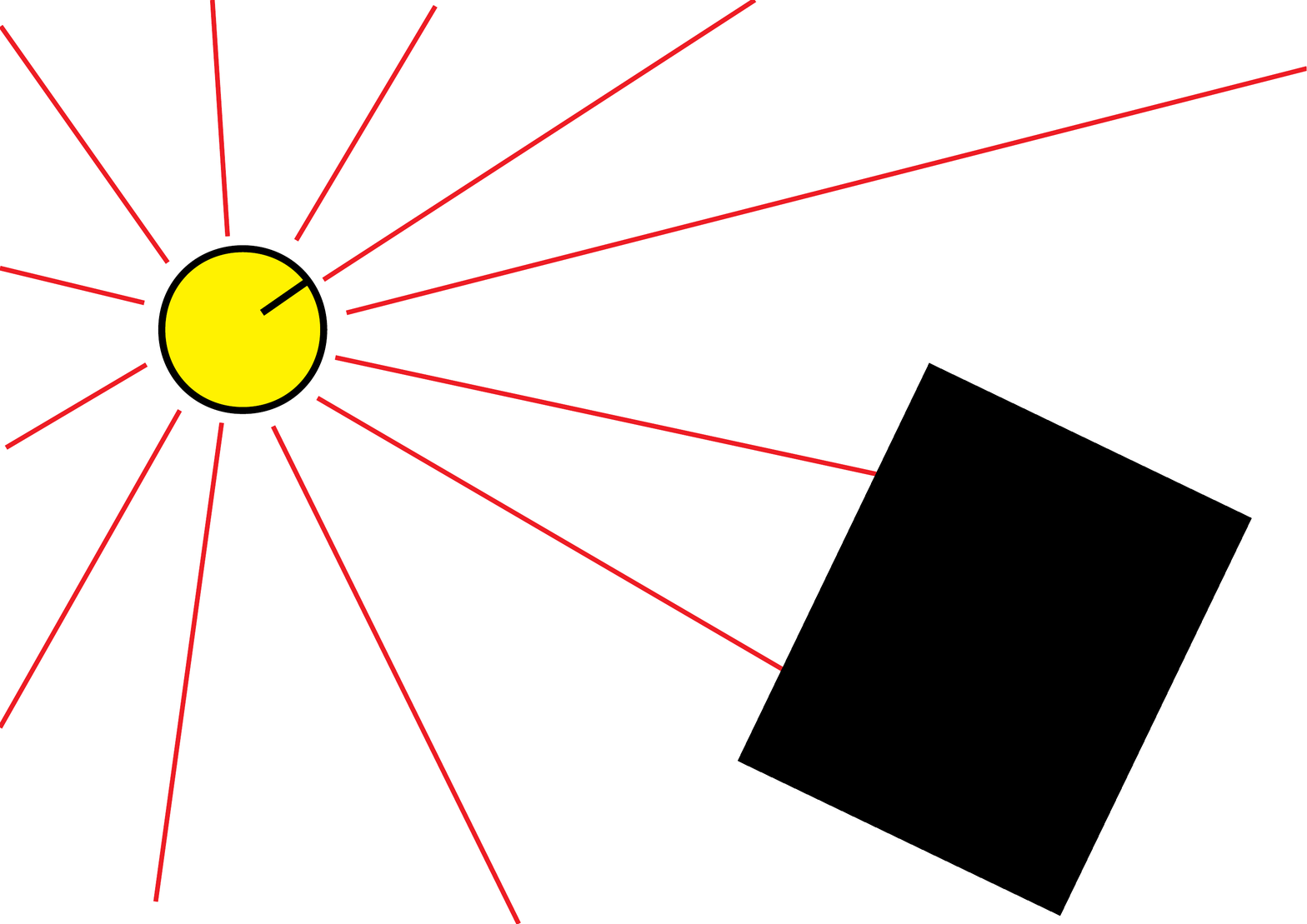}};
      \node[draw, rectangle, text centered, minimum width=1, minimum height=0.5]                    (x)    at (1.5,0)       {\includegraphics[trim=3.0cm 3.0cm 3.0cm 3.0cm, width=0.1\linewidth]{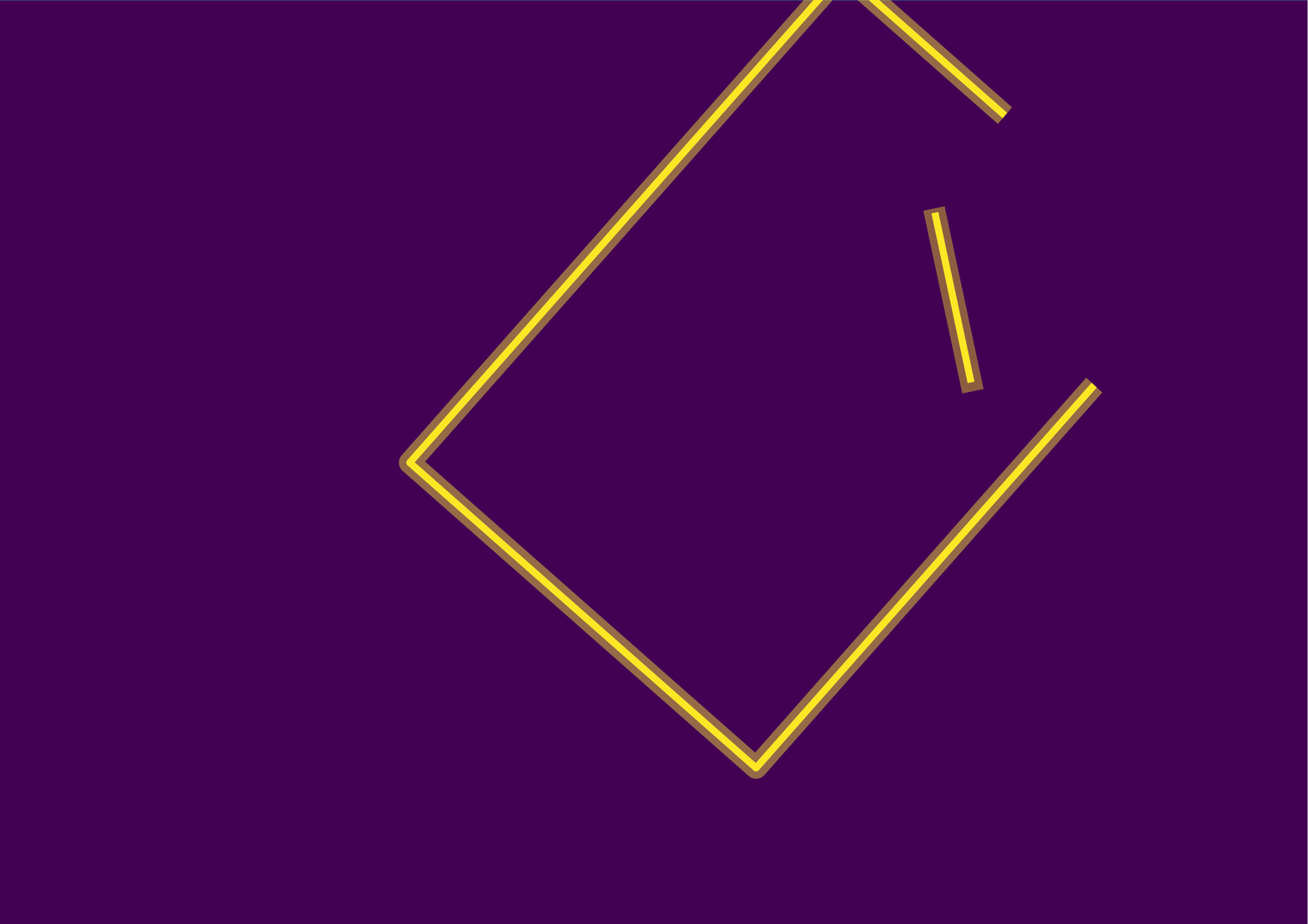}};
      \node[draw, tri, text centered, minimum width=1, minimum height=0.5]                          (e)    at (3.5,0)       {Enc.};
      \node[draw, circle, text centered, minimum width=1, minimum height=0.5]                       (z)    at (5.1,0)       {z};
      \node[draw, tri, left, text centered, minimum width=1, minimum height=0.5]                    (d)    at (6.7,0)       {Dec.};

      \node[draw, rectangle, text centered, rounded corners, minimum width=12mm, minimum height=7mm]                    (pol)    at (5.1,-1.3)       {Policy};
      \node[draw, rectangle, text centered, rounded corners, minimum width=12mm, minimum height=7mm]                    (env)    at (0.0,-1.3)       {Env.};

      \draw[->] (r) -- (x);
      \draw[->] (x) -- (e);
      \draw[->] (e) -- (z);
      \draw[->] (z) -- (d);
      \draw[->] (z) -- (pol);
      \draw[->] (pol) -- (env);
      \draw[->] (env) -- (r);
      \draw[->]  (d) -- (7.8, 0);

      \begin{scope}[on background layer]
        \node[draw=lightgray, fit=(e)(d), yshift=2mm, text height=-1em, minimum height=18mm, rounded corners, inner xsep=10, inner ysep=10, outer sep=5] (sb) {VAE};
        \node[draw=lightgray, fit=(r)(x), yshift=2mm, text height=-1em, minimum height=18mm, rounded corners, inner xsep=10, inner ysep=10, outer sep=5] (cur) {Preprocessing};
       \end{scope}
  \end{tikzpicture}

  \caption{Proposed autoencoding pipeline embedded in a RL setting. The range measurements, received from the environment, are preprocessed into the local image centered at the robot pose. The encoder compresses the image into a stochastic latent space~$z$. This is the quantity of interest and passed to the policy. The decoder reconstructs a distribution over the original image, which is required for semi-supervised optimization of the autoencoder.}
  \label{fig:pipe}
\end{figure} 

In this section, we describe each step of our autoencoding pipeline for 2D-lidar scans, which has the purpose of generating compact state representations for RL in navigation tasks. An overview is given in~\figref{fig:pipe}. At first, we introduce our novel preprocessing for autoencoding 2D-lidar scans and then present the network architecture. All neural networks are implemented and optimized using tensorflow~\cite{tensorflow}.

\subsection{Preprocessing}
Wakita~\etal~\cite{wakita18} mention that a VAE smooths sharp edges during reconstruction. We had similar observations and additionally noticed even worse behavior in cluttered areas, where the ranges of neighboring measurements underlay a lot of change. During our work, we observed that this behavior is originated in the usage of a 1D convolutional layer and significantly impacts the reconstruction performance of the autoencoding pipeline.

To overcome this, we preprocess the range measurements~$r$ into a local, egocentric, binary map, representing the occupancy in a \mbox{20\,m $\times$ 20\,m} area around the sensor. First, we normalize the data with a maximum range of 10\,m into the \mbox{interval~$\overline{r} \in [0,1]$} and set every measurement that is invalid or outside this range to zero. The normalized polar coordinates are then converted into Cartesian space and casted to indices after multiplying them with the desired map resolution. The indices are used to compute the desired binary image, referred to as local image.

Of particular note is that our local image only requires local sensor data and no global information, in contrast to the one proposed by Narasimhan~\etal~\cite{narasimhan20} or Regier~\etal~\cite{regier20}. Furthermore, this preprocessing makes the autoencoder in general invariant to the scan resolution. Therefore, different types of sensor models with different technical specification can share one jointly trained autoencoder or even contribute to the same local image at the same time when attached to the same mobile platform.

\subsection{Network Architecture}

\begin{table}
\centering
{\footnotesize
    \begin{tabularx}{0.95\linewidth}{|Y||Y|}
\hline
\textbf{Encoder} & \textbf{Decoder} \\ \hhline{|=||=|}
\textbf{Input}: $320 \times 320$ local image & \textbf{Input}: latent sample $\in \mathbb{R}^{k}$ \\ \hhline{|-||-|}
    Conv. $32 \times 3 \times 3$, stride 2 ReLU, BN & Dense, 256, ReLU \\ \hhline{|-||-|} 
    Max pool. $2 \times 2$ & Dense, $5 \times 5 \times 128$, ReLU reshaped\\ \hhline{|-||-|} 
    Conv. $32 \times 3 \times 3$, stride 2 ReLU, BN & Trans. Conv., $128 \times 3 \times 3$ stride 2, ReLU \\ \hhline{|-||-|} 
    Conv. $64 \times 3 \times 3$, stride 1 ReLU, BN & Trans. Conv., $64 \times 3 \times 3 \quad$ stride 2, ReLU \\ \hhline{|-||-|} 
    Avg. pool. $2 \times 2$ & Trans. Conv., $64 \times 3 \times 3 \quad$ stride 2, ReLU\\ \hhline{|-||-|} 
    Conv. $128 \times 3 \times 3$, stride 2 ReLU, BN & Trans. Conv., $32 \times 3 \times 3 \quad$ stride 2, ReLU \\ \hhline{|-||-|} 
    Conv. $128 \times 3 \times 3$, stride 2 ReLU, BN, flatten & Trans. Conv., $32 \times 3 \times 3 \quad$ stride 2, ReLU \\ \hhline{|-||-|} 
    Dense, 256, ReLU & Trans. Conv., $32 \times 3 \times 3 \quad$ stride 2, ReLU \\ \hhline{|-||-|}
    Dense, $2\text{k}$ & Conv. $1 \times 3 \times 3$, stride 1 \\ \hhline{|-||-|}  
    \textbf{Output}: Diag. Gaussian & \textbf{Output}: Ind. Bernoulli \\ \hline
\end{tabularx}
}
\caption{Network architecture from top to bottom.}
\label{tab:arch}
\end{table}

Our network architecture for the VAE is shown in~\tabref{tab:arch}. Each of the convolutional layers that are marked with a \textit{BN} have a preceding batch normalization layer~\cite{ioffe15}. The final layers of the encoder and decoder are distribution layers~\cite{dillon17}, which allows us to directly compute the loss function given in~\eqref{eq:elbo} and makes any further metric obsolete. For the encoder, we used a diagonal Gaussian distribution and an independent Bernoulli distribution for the decoder. Regarding the encoder, the second dense layer is used to parametrize the diagonal Gaussian distribution and therefore \mbox{has~$2\text{k}$} nodes, per latent dimension the mean value and the standard deviation.

Concluding, we have introduced a novel preprocessing as well as our network structure to encode 2D-lidar scans into a compact state representation. 
\section{Experimental Environments}
\label{sec:exp}
In this section, we present the datasets used for pretraining the autoencoding pipeline. Furthermore, we describe the navigation task on which we test our state representation. 

\subsection{VAE Datasets}
First, we shortly introduce the characteristics of all three datasets used for pretraining the VAE. The first dataset consists of a small rectangular room and acts as a simplified, simulated test case. Both side lengths are uniformly drawn from $l \in [4\,\text{m}, 8\,\text{m}]$ and a small round pole with a diameter of 0.2\,m is placed at random in the room. During simulation the robot is set to different poses and randomly drives through the environment until a collision occurs or a maximum number of steps is reached. We collected 10 trajectories per room with 250 randomly sampled rooms in total. This simulation also represents the simplistic navigation task in the next section and is therefore referred to as \textit{simple environment}, see~\figref{fig:rawImg} as an example.

\begin{figure}[t]
  \centering
  \begin{tikzpicture}
      \node[rectangle, text centered, inner sep=0mm]      (m0)    at (-2.15,0)       {\includegraphics[width=0.49\linewidth]{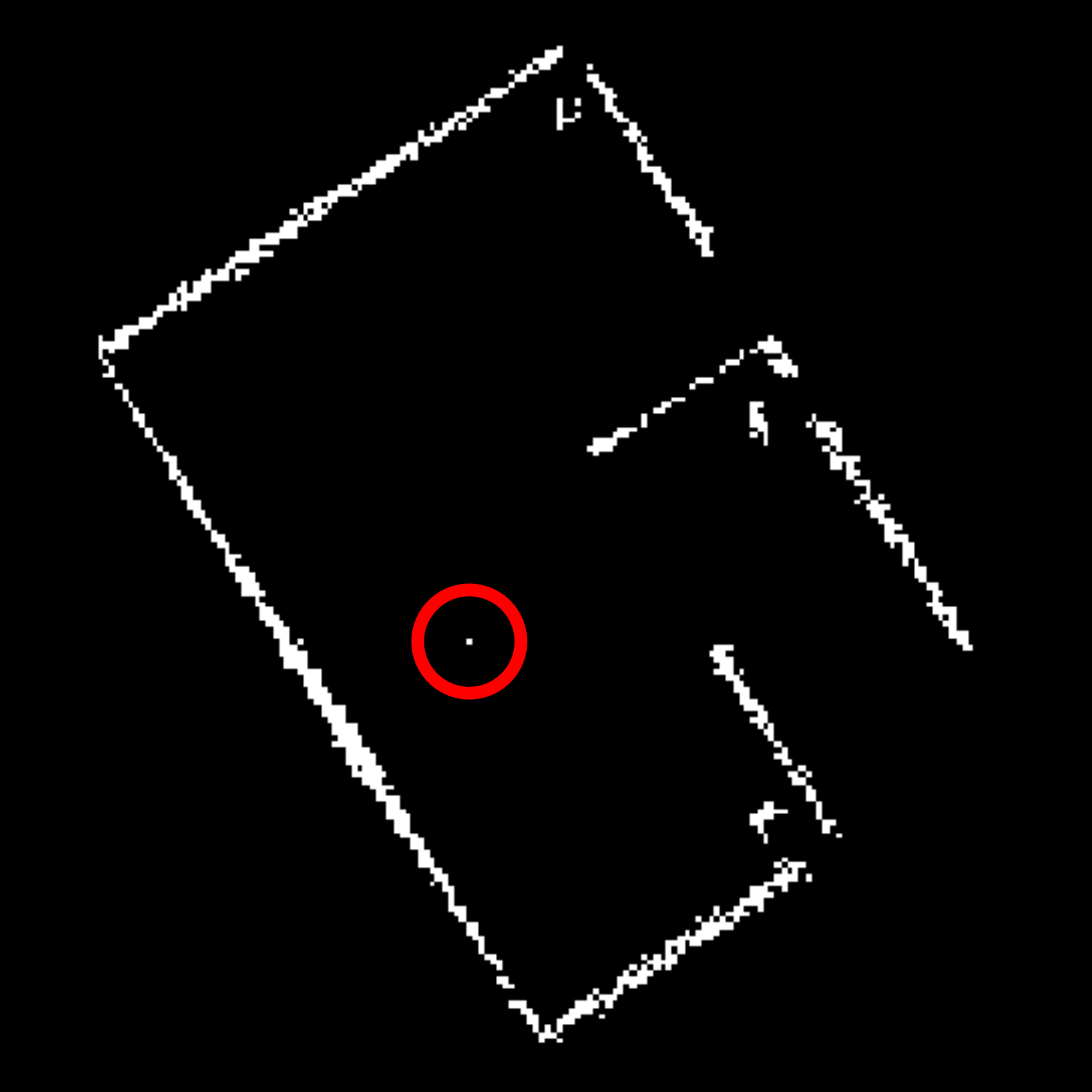}};
      \node[rectangle, text centered, inner sep=0mm]      (m1)    at (+2.15,0)       {\includegraphics[width=0.49\linewidth]{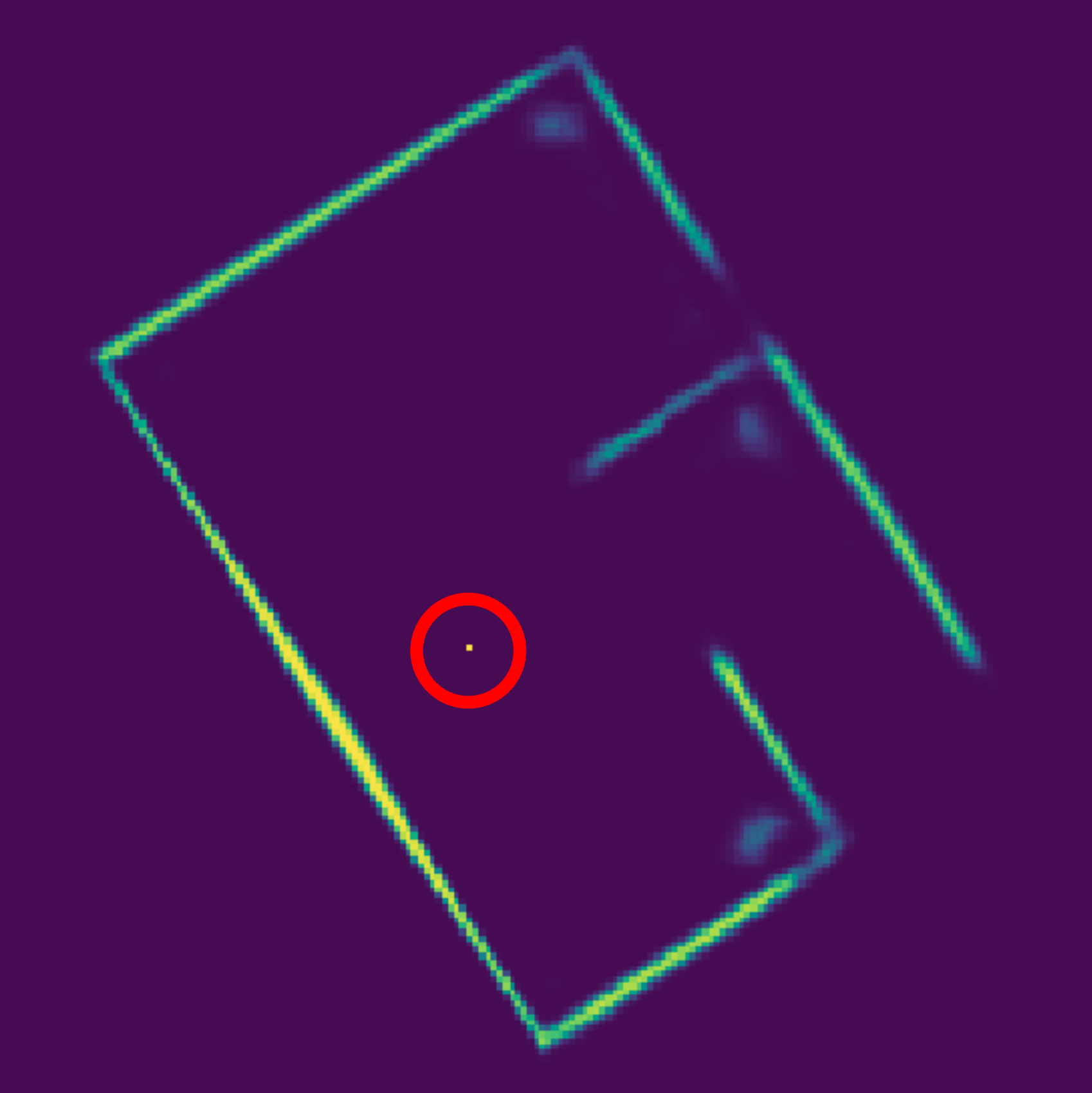}};
  \end{tikzpicture}

  \caption{A scan from the main environment, after converting it with our preprocessing~(left). On the right the reconstructed distribution returned from the generative model of the autoencoding pipeline. Brighter regions represent pixels with higher probability being occupied. As can be seen, all poles and the walls are correctly reconstructed. The robot pose is marked with a red circle.}
  \label{fig:sim_room}
\end{figure} 

The second simulated environment is more complex. We set the room size \mbox{to~$l \in [5\,\text{m}, 10\,\text{m}]$} and added additional walls, which leads to regions that cannot be observed by the lidar. We generate a total number of 100 randomly sampled rooms, where we collect in each room 250 trajectories. To simulate cluttered regions we increased the number of poles, which now can be round and square with alternating diameter~$d \in [0.1\,\text{m}, 0.4\,\text{m}]$. The likelihood of poles being placed close to each other or in corners is increased, while we limit the number of poles to 0.25 per square meter averaged across the total area in each room. We refer to this environment as \textit{main environment}. A scan and its reconstruction from this environment are visualized in~\figref{fig:sim_room}. The data are collected at a rate of 5\,Hz and simulated using pybullet~\cite{coumans19}. Noise is included by adding Gaussian blur to the true measurements and a certain percentile amount of beams is randomly invalidated.

To additionally show that our autoencoding pipeline can cope with realistic complexity outside of a simulation, we collected scans with a slamtec rplidar A3 attached to a TurtleBot in our building. We collected roughly 20,000 scans at a rate of 5\,Hz, including hallways, offices, kitchens, and labs. This dataset is referred to as \textit{real environment}, see~\figref{fig:motivation} as an example. We use for the real and simple environment a latent dimension of~$k=16$ and for the main environment a dimension of~$k=32$. Both values turned out to be the minimal value without noticing a greater reconstruction loss.

\subsection{Navigation Task}
We now shortly introduce the navigation task solved by the RL agent. It is modeled as Markov Decision Process~(MDP). The robot is originated in some state~$s_t \in \mathcal{S}$, which is provided by the autoencoding pipeline based on the currently observed lidar scan. According to the policy~$\pi$, the optimal action~$a_t \in \mathcal{A}$ is chosen. By executing this action, the agent receives a reward~$r_t$ and the next state~$s_{t+1}$, defined by the transition probability distribution~$P(s_t, a_t, s_{t+1})$. For policy optimization, we use a deterministic policy gradient method called twin delayed deep deterministic policy gradient~(TD3) introduced by Fujimoto~\etal~\cite{fujimoto18}.

As already mentioned, the environment is given by the simple environment described above and the agent is supposed to navigate towards the pole placed randomly in the rectangular room. While multiple approaches have demonstrated to generate motion commands given a desired target position~\cite{pfeiffer16},~\cite{regier20}, we want to address the proposal of intermediate target poses. This has especially the difficulty of sparse rewards, as no progress on a global path, used as input, can be considered for subrewards. Subrewards in general motivate the agent to reach its final goal and therefore experience the actual reward.

The RL agent is supposed to generate target positions relative to the robot's current position and orientation, while the actual reaching of this target pose is assumed to be achievable by a different algorithm. For our simulation, we directly place the robot on the target position and check if any collisions with walls would have occurred on the direct path. The target pose is accepted from the interval [-2.0\,m, 2.0\,m] in x and y-direction separately to limit the maximal distance. The agent is rewarded a positive signal when the robot reaches the pole within a threshold of 0.1\,m on the direct path to the proposed target position. When the robot bumps into walls, due to target poses within walls or outside of the room, the agent is penalized. In all other time steps, the agent receives no signal. An episode ends, when either the pole is reached, the robot bumps into walls, or a maximal number of relative target poses have been proposed.
\section{Results}
\label{sec:results}
In this section, we present the results of our experiments. In particular, we first show the increased reconstruction power due to the preprocessing into our local image and the effect on step edges. Afterwards, we demonstrate the effect on the downstream RL task and the general pitfall that increased reconstruction power always improves the training of the agent.
\subsection{Reconstruction Error}
\begin{figure}[t]
  \centering
 \includegraphics[width=0.75\linewidth]{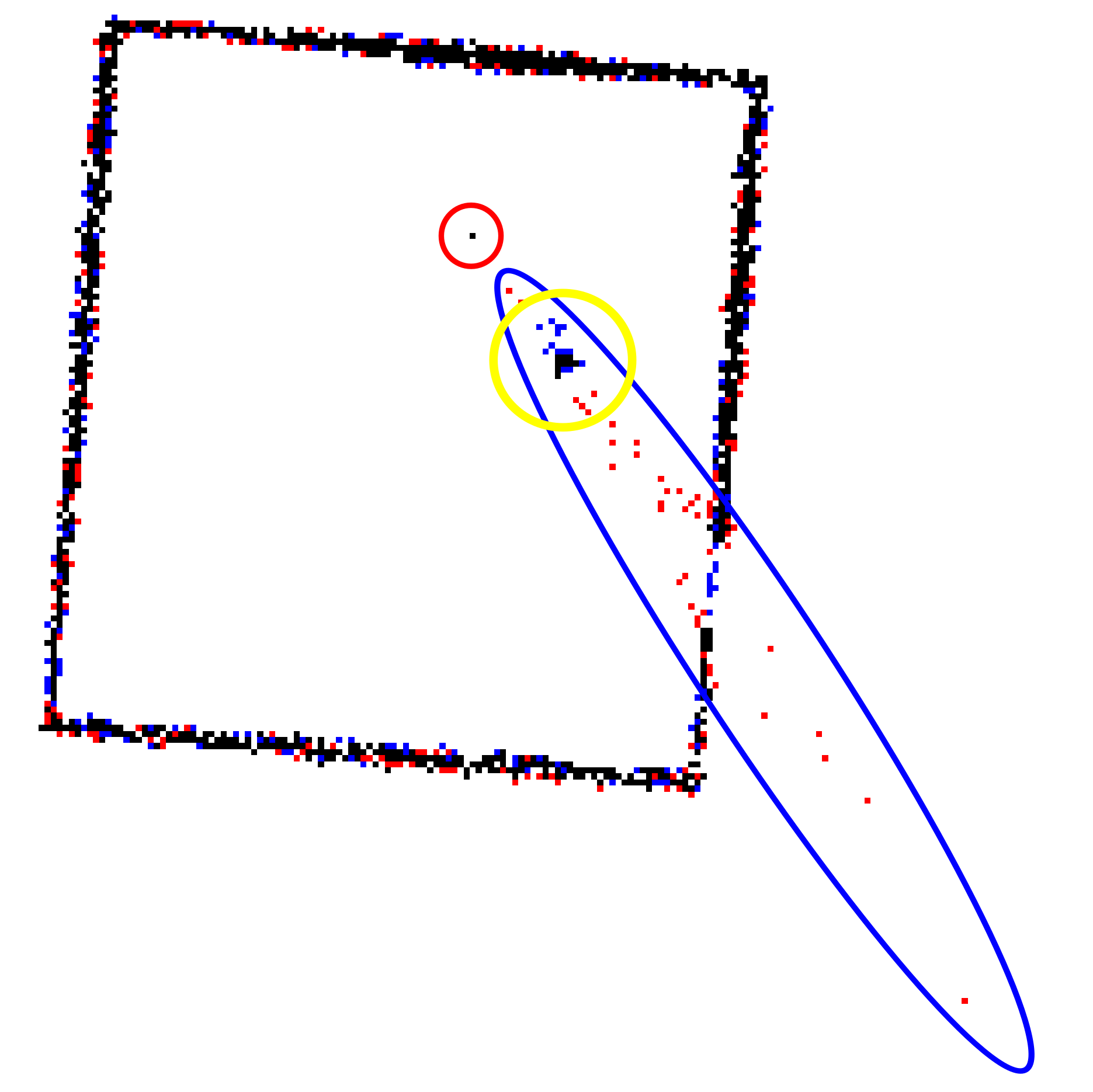}
    \caption{Reconstruction sample from the simple environment. We
      compare the autoencoding pipelines based on our local image and
      on raw measurements. False positives are marked in color, blue
      ones are based on the local image and red ones on raw
      measurements. The robot pose is highlighted with a red circle,
      which is not in the center as we cropped the free space. The
      pole is highlighted with a yellow circle. As can be seen, the
      undesired blurred edge occurs only with the raw measurements (as
      highlighted with a blue ellipse).}
  \label{fig:rawImg}
\end{figure}

In this section, we are showing the improved reconstruction power of the autoencoding pipeline due to our preprocessing. As baseline, we extend the network architecture of the encoder proposed by Pfeiffer~\etal~\cite{pfeiffer16} by an generative model. The generative model outputs a Gaussian distribution over the distance measurement for each lidar beam in one scan. The structure of the network is similar to the one proposed in~\tabref{tab:arch}, but we adapted the filter numbers, kernel sizes, and strides to meet the encoder and data structure. In contrast to the original work, we used circular padding~\cite{schubert19} in the 1D-convolutional layer. Furthermore, we normalized the raw lidar scan and replaced invalid measurements with the average of neighboring beams, as setting them to zero lead to worse reconstruction.

For the local image, we used a resolution of 320 pixel, which corresponds to 6.25\,cm per pixel in a 20\,m $\times$ 20\,m area. In \figref{fig:rawImg}, we compare the two methods on a 2D-lidar scan from the simple environment. As can be seen, learning on raw measurements blurs sharp edges, which increases false positives~(colored pixels). False positives represent pixels that have been considered as occupied while the actual scan considers this pixel as free. The reconstructed image is either computed based on a sample from the reconstructed raw measurements~(baseline) or directly sampled from the generative model of our autoencoding pipeline. 

For further comparison, we train both autoencoder pipelines in the main environment and on the recorded 2D-lidar scans from our building. To quantify the performance, we average the number of false positives and false negatives per scan over each of the datasets. The false positives~(FP) are computed as described above, while the false negatives~(FN) represent cells that are reconstructed as free but are actually occupied. Both quantities are based on sampling from the generative model. Additionally we consider the mean squared error~(MSE), computed based on the local image as the raw measurements can be converted without loss in accuracy. If the raw measurements are perfectly reconstructed the computed image base on those is also identical to the image computed on the 2D-lidar scan. For the MSE we explicitly take the expectation of the generative model from both pipelines instead of sampling from the distribution. Additionally, by using the expectation, we take into account that the reconstructed local image based on our pipeline returns the likelihood of a pixel being occupied.

\begin{table}
\centering
{\footnotesize
    \begin{tabularx}{0.9\linewidth}{|c|c|Y|Y|Y|}
\hline
        Environment & Method & FP & FN & MSE \\ \hline \hline
        \multirow{2}{*}{Simple} & our & \textbf{240.0} & 242.13 & \textbf{240.28} \\ \cline{2-5}
         & raw & 266.9 & 238.11 & 474.85 \\ \hline
        \multirow{2}{*}{Main} & our & \textbf{265.9} & 305.18 & \textbf{299.06} \\ \cline{2-5}
         & raw & 290.33 & 305.94 & 621.98 \\ \hline
        \multirow{2}{*}{Real} & our & \textbf{158.53} & \textbf{159.71} & \textbf{164.27} \\ \cline{2-5}
         & raw & 622.77 & 219.55 & 532.54 \\
\hline
\end{tabularx}
}
\caption{Error metrics comparing the preprocessing methods for each of the datasets. Each value is averaged over the entire dataset. The local image is indicated by \textit{our}, the raw measurements based on Pfeiffer~\etal~\cite{pfeiffer16} by \textit{raw}. We compute false positives~(FP) and false negatives~(FN) over the image compared to the actual scan. To measure the overall reconstruction capability, we compute the mean squared error~(MSE) for the expectation of the generative model. Our pipeline based on the local image clearly outperforms the one based on raw range measurements. After Welch's t-test~\cite{welch47} the local image leads to a significantly smaller MSE on all datasets~(for a level of 0.01).}
\label{tab:rawImg}
\end{table}

All metrics are listed in~\tabref{tab:rawImg}. The table shows that the local image increases the reconstruction power across all datasets. Especially the MSE is significantly lower after Welch's t-test~\cite{welch47}~(for a level of 0.01) in comparison to the pipeline based on Pfeiffer~etal~\cite{pfeiffer16}. While the FN are mostly in the same magnitude, the FP on the other side are lower when our preprocessing is applied across all three datasets. This strengthens the assumption of the blurring problem in~\figref{fig:rawImg}, as the pixels belonging to a step edge are captured by the FPs.

To show the transferability, we physically rebuilt the simple environment and tested our proposed pipeline without retraining it on the real scans. While for real-world applications the autoencoder as well as the RL agent should always be trained on the real data, the transferability could be interesting when quickly testing the virtually trained agent in a real-world scenario. The MSE of our autoencoding pipeline trained in simulation and tested on the physically rebuild environment is 244.86, which is in the same magnitude as the MSE for the simple environment in~\tabref{tab:rawImg}.
\subsection{Reinforcement Learning}
\begin{figure}[t]
  \centering
 \includegraphics[width=1.0\linewidth]{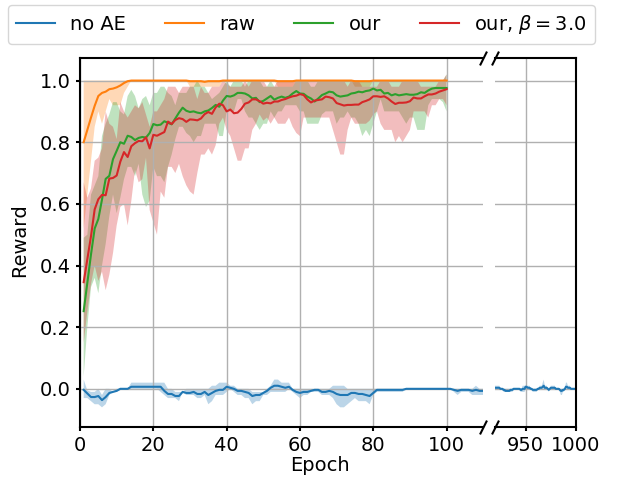}
 \caption{The reward reached by the agent over the trained epochs. We averaged the results over 10 trials for each RL setup, besides the agent not based on a pretrained autoencoder that is only trained three times. The marked background represents the minimal and maximal value in each epoch. We evaluated at the end of each epoch and averaged the received reward over 20 episodes. The agent without a pretrained autoencoder is represented by the blue line and basically does not train at all. The orange line uses the autoencoder based on Pfeiffer~\etal~\cite{pfeiffer16}. Our autoencoding pipeline is represented by the green line and with an increased $\beta$ by the red line. It is clearly visible that the improved reconstruction power leads to a decrease in the resulting performance of the RL agent in terms of collected reward. This is surprising, as improving the preprocessing pipeline should also result in an increase in performance of the RL task.}
  \label{fig:rl}
\end{figure}
In this section, we apply the pretrained autoencoders to our introduced RL task for robot navigation. The results are visualized in~\figref{fig:rl}. All agents consist of two dense layers, each with 128 nodes. The agent without pretrained autoencoder~(blue line) has an additional untrained preprocessing pipeline identical to Pfeiffer~\etal~\cite{pfeiffer16}. As can be seen, the agent does not learn at all. The amount of parameters is too big to be optimized within 1000 epochs, which roughly corresponds to three days of training. Especially the fact that the reward is only rarely received amplifies the slow convergence speed in this specific RL task.

In contrast, all of the agents based on a pretrained autoencoder solve the task completely. However, the increased reconstruction power of our autoencoding pipeline slows down the training and even leads to random reward drops after the task has been solved. This means that the agent "forgets" the optimal policy for multiple epochs in a row, before again acting optimal. Therefore, as we average each RL setup over 10 trials, the mean curve in~\figref{fig:rl} of the agent based on our pipeline~(green line) is below the maximal reward of 1.0. Higgins~\etal~\cite{higgins17} introduced $\beta$-VAE~\cite{higgins17} to simplify the latent space while keeping the same amount of information. We trained different values for $\beta$ ranging up to 5.0, where $\beta=3.0$ leads to the best results. However, we found no benefit in increasing $\beta$ in our particular situation~(red line) in terms of collected reward or training speed.

Even tough our autoencoding pipeline significantly improved the reconstruction capabilities in comparison to a pipeline based on Pfeiffer~\etal~\cite{pfeiffer16}, it turns out to that the RL agent using the compressed state representation does not benefit from the improvement in terms of collecting higher rewards. However, a simple guideline can help to avoid the exposed pitfall. Thus, when improving autoencoding pipelines for a RL downstream task a pipeline-specific metric is not sufficient and therefore the performance of the RL agent needs to be taken into account concurrently.
\section{Conclusion}
\label{sec:conclusion}
In this paper, we presented an autoencoding pipeline for 2D-lidar data and proposed a novel preprocessing that significantly improved the reconstruction power in comparison to autoencoding the raw sensor data. We demonstrated the compression capabilities on simulated and real-world datasets. Furthermore, we used these pretrained autoencoders to train a RL agent based on their latent space to solve a navigation task. Even though our autoencoding pipeline has increased reconstruction capabilities, the performance of the downstream RL agent decreases. To avoid this pitfall, the input pipeline for a RL agent cannot be optimized detached from the actual RL task. When improving the input pipeline a purely pipeline-specific metric is not sufficient and also needs to be validated by the actual performance of the RL agent using the learned state representation.

\bibliographystyle{IEEEtran}

\bibliography{bibliography}

\end{document}